\documentclass[conference]{IEEEtran}
\IEEEoverridecommandlockouts
\usepackage{cite}
\usepackage{amsmath,amssymb,amsfonts}
\usepackage{algorithmic}
\usepackage{graphicx}
\usepackage{textcomp}
\usepackage{xcolor}
\usepackage{diagbox}
\usepackage{bbm} 
\def\BibTeX{{\rm B\kern-.05em{\sc i\kern-.025em b}\kern-.08em
    T\kern-.1667em\lower.7ex\hbox{E}\kern-.125emX}}

\usepackage{cite}

\begin{document}

\title{Neural network interpretability for forecasting of aggregated renewable generation}


\author{
\IEEEauthorblockN{Yucun~Lu, Ilgiz~Murzakhanov, Spyros~Chatzivasileiadis}
\IEEEauthorblockA{Center for Electric Power and Energy \\
Technical University of Denmark\\
Kgs. Lyngby, Denmark\\
luyucun003@gmail.com, \{ilgmu, spchatz\}@elektro.dtu.dk}
}

\maketitle

\begin{abstract}

With the rapid growth of renewable energy, lots of small photovoltaic (PV) prosumers emerge. Due to the uncertainty of solar power generation, there is a need for aggregated prosumers to predict solar power generation and whether solar power generation will be larger than load.
This paper presents two interpretable neural networks to solve the problem: one binary classification neural network and one regression neural network. The neural networks are built using TensorFlow. The global feature importance and local feature contributions are examined by three gradient-based methods: Integrated Gradients, Expected Gradients, and DeepLIFT. Moreover, we detect abnormal cases when predictions might fail by estimating the prediction uncertainty using Bayesian neural networks. Neural networks, which are interpreted by the gradient-based methods and complemented with uncertainty estimation, provide robust and explainable forecasting for decision-makers.
\end{abstract}


\section{Introduction}


Small renewable energy producers, such as photovoltaic (PV) solar panels, are relatively new at the distribution grid level. For example, in Australia, the government used to incentivize small households to install solar panels by offering generous feed-in tariff schemes \cite{guardian}. On top of covering part of their own consumption, the owners of small PVs were paid for injecting the power surplus to the grid. However, with greater penetration of renewable generation at the distribution grid level, PV inverters started to turn off at peak generation hours due to voltage rising in the network \cite{guardian}. 
While there can be several solutions to tackle this problem, the one requiring the lowest investments is the forecasting of aggregated renewable generation and aggregated consumption. From the conducted forecast, the decision-makers may conclude about generation and consumption balance and decide to curtail part of solar production. In this work, we propose methods that can (i) interpret the forecasting techniques, and (ii) can assess how certain a predicted value is. 

\subsection{Literature Review}
Several research efforts have been dedicated to building models to predict solar power generation and load with the focus on either reducing forecasting errors or improving computation efficiency. The traditional methods for load and solar power predictions are statistical algorithms, such as polynomial regression and auto-regressive and moving average model (ARMA). In \cite{LR}, the authors present a linear regression model to predict solar radiation using temperature data. In \cite{ARMA}, the authors propose an hourly ARMA model to predict power output from a PV panel. Recently, a great number of researchers work on the development and improvement of several machine learning algorithms in solar power forecasting. The algorithms include support vector regression (SVR), gradient boosted regression tree (GBRT), and artificial neural network (ANN). In \cite{SVR}, the researchers present an SVR model for solar power forecasts on a rolling basis for 24 hours ahead. In \cite{GBRT}, a GBRT model is proposed using historical power generation data and relevant meteorological variables. The study in \cite{ANN} suggests ANN for producing solar power forecast. The study in \cite{LSTM} and \cite{CNN} applies long short-term memory (LSTM) and convolutional neural networks with LSTM (CNNs-LSTM) respectively to provide accurate forecast results. 

While machine learning algorithms provide promising results in load and PV forecasting, they lack interpretability. As a result, explainable artificial intelligence (XAI) has become an emerging research direction, which addresses this problem and helps understanding why and how these models make predictions. Currently, only a few works exist in this field. In \cite{x1}, the authors propose a unified clustering-based prediction framework with two tree-based algorithms and use the SHapley Additive exPlanations (SHAP) XAI tool to interpret the model. In \cite{x2}, the authors suggest an interpretable forecasting model using the XGBoost algorithm and ELI5 XAI tool. In \cite{x3}, the authors apply several XAI tools: LIME, SHAP, and ELI to interpret a random forest model.

\subsection{Main contributions}
The main contributions of this paper are the following:
\begin{itemize}
    \item We conduct local and global feature importance analysis for the designed classification and regression neural networks. From the feature analysis, we discover several interesting dependencies between the features and the outputs, which match with the physical phenomena behind the PV generation process. 
    \item We implement three gradient-based methods for interpreting the designed models and find the main reasons why the neural networks fail. We also give recommendations on building robust neural networks.  
    \item We design a Bayesian neural network to estimate the aleatoric and epistemic uncertainties of the solar power forecast models, which was not done before. As a result, we detect the abnormal PV generation hours, when our models might fail, and consider these hours more carefully. 
\end{itemize}
 
\subsection{Outline}
The remainder of this paper is organized as follows. First, the description of the proposed methodology is given in Section~\ref{sec:methodology}. Numerical results with the following analysis are provided in Section~\ref{sec:NumRes}. Finally, Section~\ref{sec:conclusion} concludes the paper and proposes future directions.

\section{Proposed Methodology}\label{sec:methodology}
Next, we describe two models, which predict whether solar power will exceed load in the next 15-60 minutes. Depending on the models' output, a decision on curtailing part of the PV generation is made. 

The first model is the regression neural network, which predicts solar power generation. By comparing it with the predicted load, we can conclude whether solar power will be larger than load. Here we use the previous hour's load as the predicted load because the load does not vary a lot in one hour compared to the solar power generation. As a result, our prediction is largely dependent on the accuracy of solar power prediction. The advantage of the regression model is that we can predict the exact amount of PV generation for curtailment.

The second model is a binary classification neural network that directly predicts whether solar power generation will be larger than load. If the classification model outputs 1, it means that solar power generation will exceed load; and vice versa if the model outputs 0. This model considers both effects of solar power generation and load. The classification model is more robust than the regression one and it can effectively avoid unwanted false-positive cases.

Implementation of both classification and regression models allows us to assess how the interpretability methods perform on the different types of neural networks. We deploy several attribution methods to explain both models by interpreting feature importance globally and locally. 
Thus, through neural network interpretability, not only do we identify the most influential features in making predictions, but also show how and why neural networks make certain predictions in individual cases. This gives unique insights into the models and can support informed decisions for the human operator.

Additionally, for the regression model of solar power prediction, aleatoric uncertainty and epistemic uncertainty are estimated with Bayesian neural networks. The interpretation of neural networks can tell us why the model predicts certain results, while the uncertainty can tell us how sure the model is about its predictions. Combining interpretation with prediction uncertainty results in more transparent, convincing, and robust decisions.

\subsection{Interpretability}
To interpret the neural networks and find the most influential features in predictions, we apply attribution methods. Attribution methods assign an attribution value, sometimes called ``contribution'' or ``relevance'', to each input feature of a neural network, which can help to determine how different features contribute to the model's output~\cite{attribution}. By looking into the attribution values of different features, we can provide reasonable explanations of how and why the prediction is made. There are five main gradient-based attribution methods: Gradient, Gradient$*$Input, Integrated Gradients, Expected Gradients, and Deep Learning Important FeaTures (DeepLIFT). Each of these methods has advantages and shortcomings, which we discuss further.

Historically, one of the first attribution methods adapted in the deep learning domain is Gradient ~\cite{gradient}. Since the gradient of the network output measures the instantaneous rate of change of model output with respect to one specific feature, it can naturally be a candidate for attribution values. Another straightforward method is Gradient$*$Input which is more commonly used when we are concerned with the marginal effect of a feature on the output. This attribution can be computed by multiplying the gradient of the model output with respect to the input itself \cite{gradient_based_method}. Unfortunately, Gradient and Gradient$*$Input methods have many unsatisfying shortcomings. One of the worst problems is called saturation:  the gradients of input features might only have small magnitudes around a sample even if the network depends heavily on those features~\cite{saturation}. The problem is common when the model output flattens after some features get to a certain magnitude. The saturation problem is significant because it might lead us to overlook some influential features but focus on less relevant features, thus yielding the wrong interpretation of networks. 

More advanced methods as DeepLIFT, Integrated Gradients, and Expected Gradients, are free of the saturation problem. As a result, we implement these three attribution methods in our work. DeepLIFT is a method for decomposing the output of the network on input features by back-propagating the contributions of all neurons in the network to every input feature \cite{deeplift}. To be more specific, DeepLIFT compares the activation of each neuron $x$ to its ``reference'' (also called baseline) activation $\bar{x}$ and assigns contribution scores according to the difference. The baseline is defined by the user and often chosen to be zero. The method of Integrated Gradients is similar to the Gradient method in that it also computes the partial derivatives of the output with respect to each input feature \cite{ig}. However, instead of computing one single gradient, Integrated Gradients computes several instantaneous gradients on the straight-line path from the baseline to the observation being explained and averages them. The method of Expected Gradients is an extension of Integrated Gradients, designed to remove baseline ambiguity \cite{eg}. To make the baseline uninformative, this method uses different baselines and takes an average over them. All three methods satisfy one desirable property called completeness that the attributions sum up to the target output minus the target output evaluated at the baseline ~\cite{gradient}. An overview of implemented gradient-based attribution methods is given in Table~\ref{tab:attribution}.

\begin{table*}[htbp]
    \caption{Implemented gradient-based attribution methods}
    \begin{center}
    \begin{tabular}{c|c}
    \hline
       \textbf{Method}  &   \textbf{Attribution}\\
       \hline
        Integrated Gradients & $ \displaystyle{\left. (x_i-\bar{x_i})\cdot  \int_{\alpha=0}^1 \frac{\partial{y_c(\tilde{x})}}{\partial{\tilde{x_i}}} \right|_{\tilde{x}=\bar{x}+\alpha(x-\bar{x})} d\alpha }$ \\
        Expected Gradients & $ \displaystyle{\int_{\bar{x}}\left(\left. (x_i-\bar{x_i})\cdot  \int_{\alpha=0}^1 \frac{\partial{y_c(\tilde{x})}}{\partial{\tilde{x_i}}} \right|_{\tilde{x}=\bar{x}+\alpha(x-\bar{x})} d\alpha \right)} p_D(\bar{x})d\bar{x}$\\
        DeepLIFT & $\displaystyle{(x-\bar{x_i})}\cdot \frac{\partial{^g y_c(x)}}{\partial{x_i}}, g = \frac{\sigma(z)-\sigma(\bar{z})}{z - \bar{z}}$\\
        \hline
    \end{tabular}
        \end{center}
    \label{tab:attribution}
\end{table*}

\subsection{Uncertainty quantification}
\label{sc:uq}
The knowledge of uncertainty is fundamental to the development of robust and interpretable machine learning techniques. The uncertainty associated with machine learning models can be broadly classified into two types: aleatoric uncertainty and epistemic uncertainty. The overall uncertainty of any model is a combination of the above two types of uncertainty. 

Aleatoric uncertainty refers to the irreducible error of the uncertainty which means the error cannot be reduced by choosing a better model because the uncertainty originates from the non-deterministic nature of the sought input/output dependency. Aleatoric uncertainty captures the uncertainty concerning information that our data cannot explain \cite{Uncertainty}. For example, in lab experiments, even when all input values are similar, the values measured after multiple trials are never the same. This type of uncertainty is a typical aleatoric uncertainty, while uncertainty due to a lack of knowledge about the perfect predictor, for example, caused by uncertainty about the parameters of a model, is called epistemic uncertainty. Epistemic uncertainty captures our limited understanding of the real-world process for which we are building the model, as we are not able to capture all the input features that affect the target variable \cite{Uncertainty}. In principle, epistemic uncertainty is a reducible error which means this uncertainty can be reduced with more knowledge about the process.

In this paper, we apply Bayesian neural networks to estimate the two types of uncertainties. Bayesian neural networks are similar to normal neural networks except that instead of having a model parameterized by its point weights, each weight of the Bayesian neural network now has a probability distribution with a mean and variance which is tuned during the training process. For each point, epistemic uncertainty is modeled by Monte Carlo sampling. In Bayesian neural networks, there is a prior distribution over the model’s weight. As a result, each time when we apply the network to make predictions, we sample the model weight from distributions. In other words, every time we run the model, we would have different model weights and thus a different predicted output. After we run the model hundreds of times, the standard deviation of model predictions can be estimated as epistemic uncertainty. The estimation of aleatoric uncertainty is much easier. Here, apart from the prior distribution put over the model's weights, we also place a distribution over the output of the model~\cite{bnn_uncertainty}. Thus, the Bayesian neural network can directly output a distribution with mean and variance. The mean value is the predicted value while the standard deviation is our estimated aleatoric uncertainty. 

\section{Numerical Results}\label{sec:NumRes}
In this section, we describe the used dataset, introduce results for the designed neural networks, and provide analysis on interpretability and uncertainty quantification. All code used for the analysis in this section is available at the GitHub repository\cite{github}.

\subsection{Dataset}
For numerical simulations, we use the data set from Presumed Open Data: Data Science Challenge, which is provided by Western Power Distribution energy data hub \cite{dataset}. The load data is extracted from the Stentaway Primary substation near Plymouth, on the south coast of the UK. The solar PV generation data is from a solar farm in Devon, UK, which is not too far from the Stentaway substation. The original load and solar power data consist of half-hourly average power values from November 2017 to July 2020. Besides, hourly irradiance forecast and hourly surface temperature forecast data from January 2015 to July 2020 have been extracted from six different sites which are close to Devon but in different directions.

We follow standard data preprocessing procedures: data cleaning, outlier detection, and data imputation. During this process, we only retain samples from 7:00 - 18:00 in each day because other hours have zero or close to zero solar power generation throughout the whole year. The final data set consists of 10623 hourly instances from November 2017 to July 2020 with 7 different attributes.

A description of all data attributes and their type is presented in Table \ref{tab:feature}. As mentioned before, we have hourly irradiance forecast data and hourly surface temperature forecast data of six sites. However, in the neural network training, we choose forecast data of only one site as input features due to the found high co-linearity in the data. Note that we consider load and generation values for an hour, as a result, use $kWh$ units further.

Note that during model training, the hour index has been encoded using a one-hot (also called ‘one-of-K’ or ‘dummy’) encoding scheme which creates new binary columns to indicate the presence of each hour from the original data. One-hot encoding can make the representation of categorical data more expressive and ensure that the neural network does not assume that higher numbers (by usual integer encoding) are more important.

\begin{table}[htbp]
\scriptsize
    \caption{Description of attributes in data set}
    \centering
    \begin{tabular}{ccc}
    \hline
         Feature & Description & Type\\
         \hline
         Index & Hour index & Categorical\\
         STemp & Surface temperature forecast ($^0C$) &  Float\\
         Irra & Irradiance forecast ($W/m^2$) &  Float\\
         PTemp & Solar panel temperature in last hour ($^0C$) &  Float\\
         HPow& Previous hour's solar power ($KWh$) &  Float\\
         DPow&  Solar power at the same hour at previous day ($KWh$)&  Float\\
         HLoad&  Previous hour's load ($KWh$) &  Float\\
         \hline
    \end{tabular}
    \label{tab:feature}
\end{table}

\subsection{Neural network models}
Conventionally, the data set is divided into training and test sets. 10\% of the data are randomly shuffled for testing and the rest are used for training. The motivation behind random shuffling is that the performance of the neural network depends on how many previously unseen data samples are contained in the test set. By randomly shuffling the data we guarantee that the model trains on as many data samples as possible while it avoids overfitting. When randomly shuffling the data set, a common random seed is used so that we can compare and validate the model with a consistent train/test split. Note that the training and testing sets are the same for the regression and classification models.

\subsubsection{Classification model}
We build a neural network of three hidden layers with 50, 30, 10 neurons, and the ReLu activation function. The output layer has one neuron and the activation function is sigmoid so that when the output is larger than 0.5, it predicts that solar power is larger than load. We choose binary cross-entropy as the loss function and Adam as the optimizer. For the classification problem, we use the accuracy metric to evaluate the model. The test accuracy of the model is 91.6\%. 



\subsubsection{Regression model}
For the solar power forecast regression problem, only the first 6 input features listed in Table \ref{tab:feature} are used. A neural network with the same structure as for the classification problem is built for the regression problem. The only difference is that now the output layer is one neuron with the sigmoid activation function combined with a lambda layer whose function is to do simple mathematical operations on the previous layer without adding more trainable weights. Here, the lambda layer is used to multiply the sigmoid output from the last layer with the solar power capacity. Since the sigmoid activation function will output from 0 to 1, when it is multiplied by the capacity of solar PV, the predictions are limited within the range from 0 to solar power capacity of 2000 kW. Thus, the neural network will not output some implausible and unpractical values like negative power generation. We select the mean squared error (MSE) as the loss function and Adam as the optimizer. The root mean square error (RMSE) metric is used to evaluate the model, which can provide a global error measure during the entire forecasting period \cite{error}. The RMSE of the solar power forecast model is around 144 kWh.

\subsection{Interpretability}
\label{sec:interpret}

Neural networks are usually referred to as ``black-box'' models due to lack of transparency. In this section, we focus on the interpretability aspects of the neural networks: why and how the models make the prediction and which features have more or fewer contributions to the final predicted results. Next, three attribution methods are implemented to explain the predictions of two models: Integrated Gradient, Expected Gradient, and DeepLIFT.

First, we obtain a global view of model behavior and find out which features are the most influential for the models' predictions. We accomplish this goal by calculating the average magnitude of feature attributions across all data set points. We conclude that the features with a larger value have more deterministic effects on the prediction results. The most important global features for the classification and regression models are shown in Table~\ref{tab:gf_lc} and Table~\ref{tab:gf_power}, respectively. We use the extreme baseline for Integrated Gradients and DeepLIFT methods, which means that the baseline prediction value is obtained by setting all features to zero.

\begin{table}[htbp]
\scriptsize  
  \centering
  \caption{Global feature importance of the classification model}
    \begin{tabular}{l|cccccc}
    \hline
          & \multicolumn{5}{c}{Feature importance} \\
    Attribution method & DPow & HPow & HLoad & Irra & PTemp & STemp\\
    \hline
    Integrated Gradients & 0.185     & 0.462    & 0.758     & 0.336     & 0.234 & 0.034\\
    Expected Gradients & 0.046    & 0.228     & 0.079     & 0.186    & 0.116 & 0.011\\
    DeepLIFT & 0.188    & 0.487     & 0.776     &  0.360    & 0.254  & 0.037\\
    \hline
    \end{tabular}
  \label{tab:gf_lc}
\end{table}

\begin{table}[htbp]
\scriptsize
  \centering
  \caption{Global feature importance of the regression model}
    \begin{tabular}{l|ccccc}
    \hline
          & \multicolumn{5}{c}{Feature importance} \\
    Attribution method  & DPow & HPow & Irra & PTemp  & STemp\\
    \hline
    Integrated Gradients     & 37.20    & 450.39     & 170.18 & 146.64 & 40.36 \\
    Expected Gradients    & 15.81 & 336.10     & 114.64     & 77.50 & 13.70 \\
    DeepLIFT   &44.22 &288.62   & 206.80     & 47.74    & 31.26\\
    \hline
    \end{tabular}
  \label{tab:gf_power}
\end{table}

For the classification model in Table~\ref{tab:gf_lc}, the most influential features are the previous hour's load (HLoad), the previous hour’s solar power (HPow), and irradiance forecast (Irra). For Integrated Gradients and DeepLIFT methods, HLoad is more influential than HPow. As for Expected Gradients, HLoad is a less important feature. The reason is that Expected Gradient uses different baselines and takes an average over them, thus the baseline of Expected Gradients is close to the average values of each feature. As discovered from the data set, solar power generation is more unstable than load. As a result, the variance of the load is much smaller than the variance of solar power generation. Thus, the influence of load is reduced during the selection of average values as the baseline. Additionally, we find that global feature importance interpreted by Integrated gradients and DeepLIFT have similar patterns: their average magnitudes of feature contributions are very close to each other. Our observations match with the previous research, where DeepLIFT is characterized as a good and fast approximation of Integrated Gradient~\cite{attribution}.

For the regression model in Table~\ref{tab:gf_power}, the previous hour’s solar power (HPow) and irradiance forecast (Irra) are the most important features for prediction. However, we observe that panel temperature (PTemp) is also an informative feature. We explain it by the physical phenomena when the high temperature of solar panels decreases their efficiency, as a result, leading to lower generation of PV panels.

\begin{figure}[htbp]
    \centering
    \includegraphics [width=0.9\linewidth]{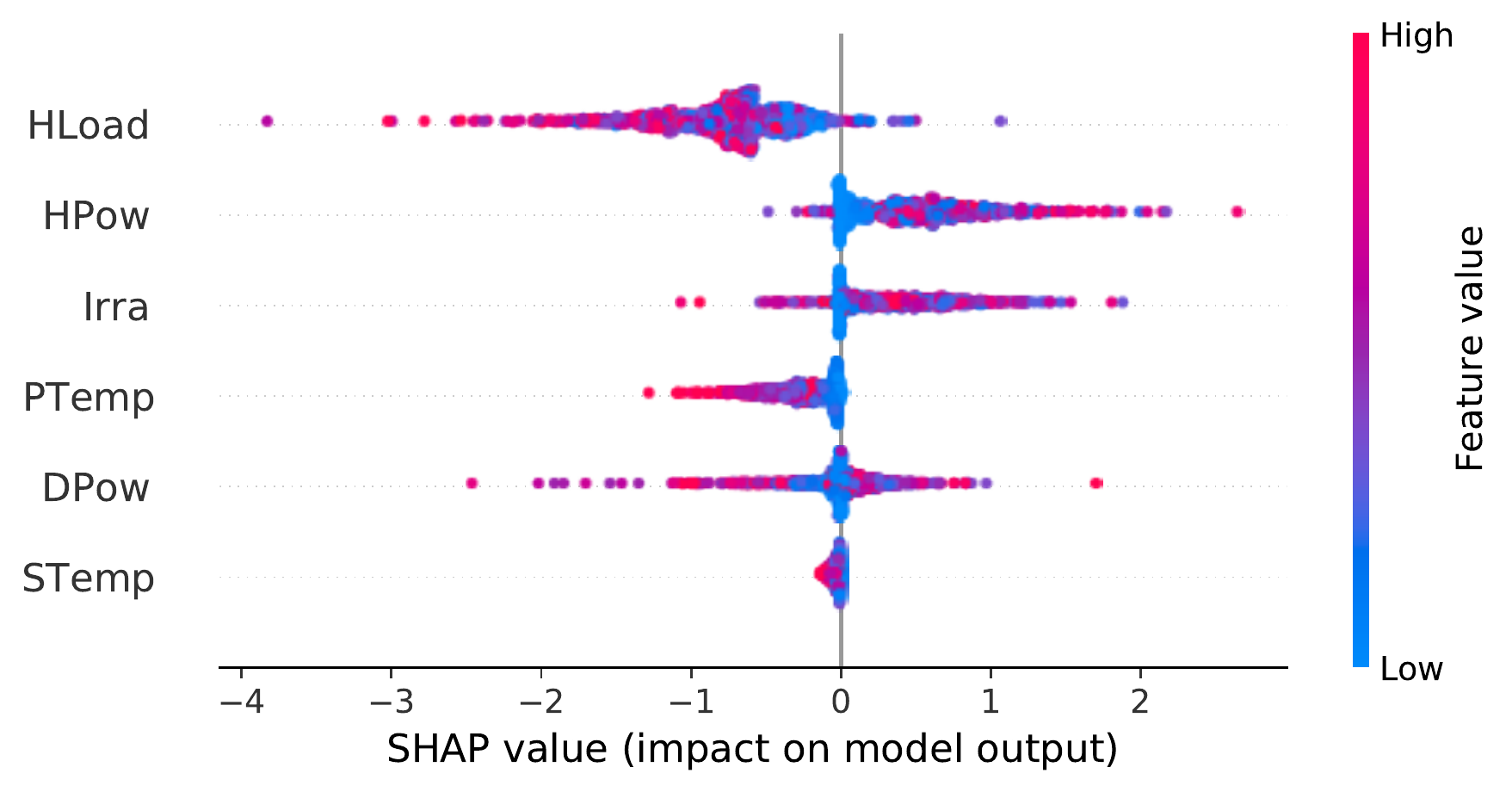}
    \caption{Summary plot of the classification model}
    \label{fig:sp_c}
\end{figure}

\begin{figure}[htbp]
    \centering
    \includegraphics [width=0.9\linewidth]{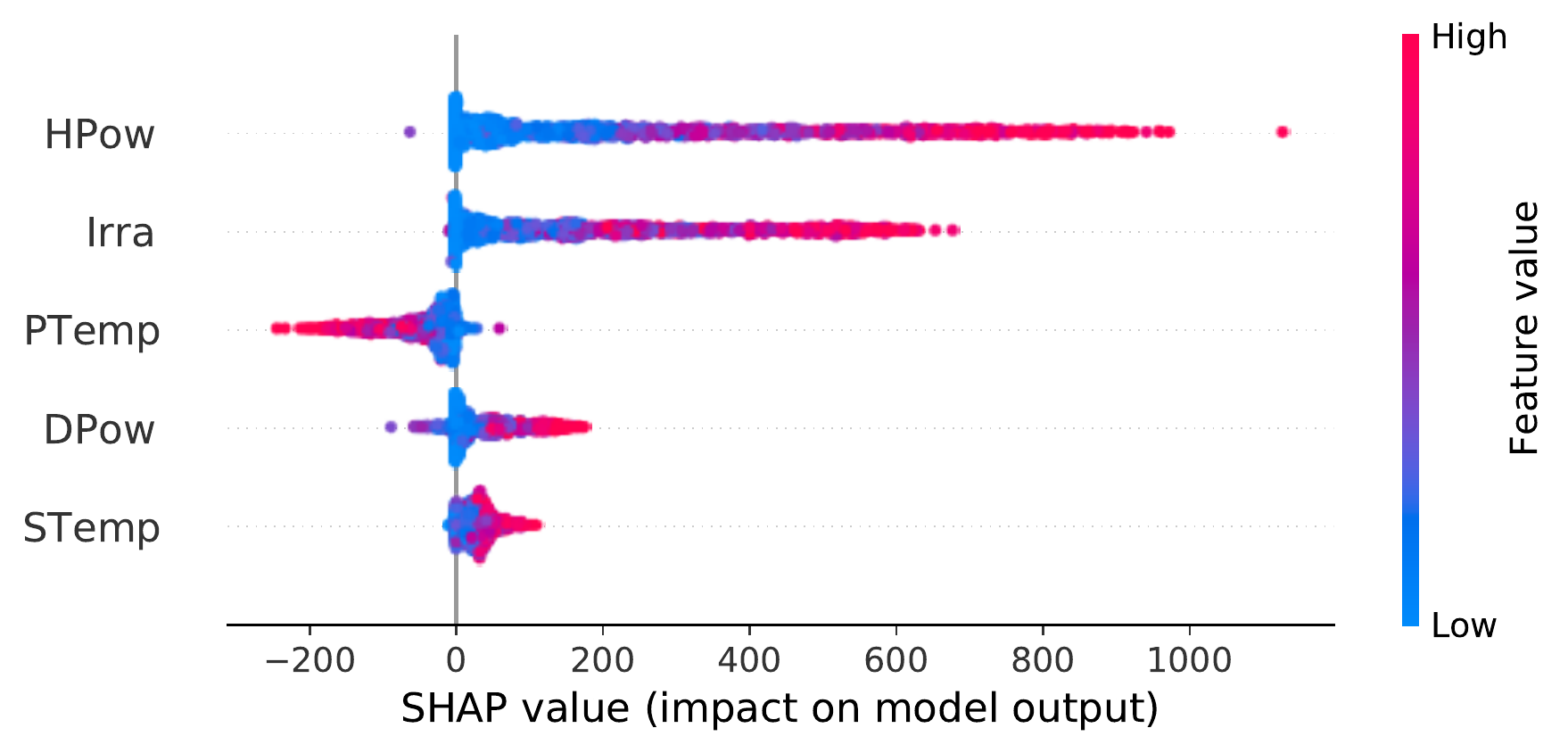}
    \caption{Summary plot of the regression model}
    \label{fig:sp_r}
\end{figure}

The aforementioned global feature analysis allows defining the most influential features for predictions. However, this analysis does not explain how exactly the features affect the predicted output. To gain more insights on that, we apply the DeepLIFT method and plot the summary plots for the classification and regression models in Fig.~\ref{fig:sp_c} and Fig.~\ref{fig:sp_r}, respectively. The summary plots show the global feature impacts on the models' output, through which we can reveal the inner workings of neural networks. On the plots, the blue dot means a point with a low feature value, while the red dot means a point with a high feature value.

For the classification model in Fig.~\ref{fig:sp_c}, HLoad generally has a negative contribution while HPow and Irra have positive impacts on model output in most cases. When HPow and Irra have high values, they tend to have large positive impacts on model output, which means the model will be more likely to predict 1 (solar power is greater than load). On the contrary, when HPow and Irra have small contributions, the model's output is more likely to be 0. For HLoad, it has negative impacts on the model output, that is, when HLoad has high values, it will promote model output to be 0, and vice versa. Both of these observations perfectly match with common-sense expectations.

For the regression model in Fig.~\ref{fig:sp_r}, the baseline features are set to be zero. As a result, Irra and HPow always have positive contributions compared to zero irradiance or zero generation in the last hour. When the regression model predicts a large solar power generation, HPow and Irra tend to have larger contributions. We also discover that PTemp has negative effects on solar power predictions. When PTemp value is high, it decreases the model output. As explained earlier, it happens due to high temperature decreases the efficiency of the solar panel. 


To get a better understanding of the role of different features contributing to the output and compare the roles those features play in each case, we examine four individual cases of local interpretability: True Positive (TP), True Negative (TN), False Negative (FN), False Positive (FP).  The attribution values of the four individual cases interpreted by DeepLIFT for the classification and regression models are presented in Table~\ref{tab:attribution_lc} and Table~\ref{tab:attribution_power}, respectively. As observations for the models are similar, we analyze Table~\ref{tab:attribution_lc} and Table~\ref{tab:attribution_power} together.

\begin{table}[htbp]
\scriptsize
  \centering
  \caption{Feature contribution to the prediction of the classification model}
    \begin{tabular}{lcccc}
    \hline
    \diagbox{Features} {Attribution} & \multicolumn{1}{c}{TP} & \multicolumn{1}{c}{TN} & \multicolumn{1}{c}{FN} & \multicolumn{1}{c}{FP} \\
    \hline
    Base value & 0.634 & 0.634 & 0.634 & 0.634 \\
    Index & -0.001 & -0.001 & -0.003 & -0.001 \\
    Stemp & -0.034 & -0.008 & -0.132 & -0.031 \\
    Irra  & 0.118 & 0     & 1.231 & 0.625 \\
    Ptemp & -0.296 & -0.04 & -0.803 & -0.296 \\
    HPow  & 0.799 & 0.154 & 0.816 & 0.425 \\
    DPow  & 0.14  & 0.018 & 0.196 & -0.047 \\
    HLoad & -0.36 & -0.755 & -1.489 & -0.318 \\
    Predicted value & 1.000 & 0.002 & 0.450 & 0.991 \\
    \hline
    \end{tabular}%
  \label{tab:attribution_lc}%
\end{table}%

\begin{table}[htbp]
\scriptsize
  \centering
  \caption{Feature contribution to the prediction of the regression model}
    \begin{tabular}{lcccc}
    \hline
    \diagbox{Features} {Attribution} & \multicolumn{1}{c}{TP} & \multicolumn{1}{c}{TN} & \multicolumn{1}{c}{FN} & \multicolumn{1}{c}{FP} \\
    \hline
    Base value & 1.32  & 1.32  & 1.32  & 1.32 \\
    Index & -68.73 & -11.13 & -42.53 & -59.76 \\
    Stemp & 45.67 & 7.57  & 45.3  & 31.93 \\
    Irra  & 534.43 & 3.09  & 226.76 & 421 \\
    Ptemp & -89.63 & -11.35 & -77.31 & -74.51 \\
    HPow  & 636.33 & 17.18 & 278.1 & 510.92 \\
    DPow  & 122.83 & 3.5   & 45.38 & 104.21 \\
    Predicted value & 1182.22 & 10.18 & 477.02 & 935.11 \\
    Real value & 1180.00 & 6.00  & 784.00 & 470.00 \\
    Predicted load & 400 & 810 & 560 & 414\\
    \hline
    \end{tabular}%
  \label{tab:attribution_power}%
\end{table}%

For True Positive (TP) case, we see that the large attribution values of Irra and Hpow lead to a high predicted solar power generation. After comparing solar generation with the predicted load, the regression model can give a prediction that solar power will exceed load. In the classification model, due to large positive attributions of Hpow and large negative attributions of HLoad, the model also gives output close to 1, which shows that the model is quite confident that solar power will exceed load. In this case, both models agree on the forecast, which is correct.  

For True Negative (TN) case, Irra and HPow have small contributions leading to a low predicted solar power generation in the regression model. Thus, when compared to the previous hour's relatively high load, it predicts that solar power is smaller than load in this hour. For the classification model, because of low Irra and HPow contributions and large negative HLoad contributions, the model also gives a quite confident prediction (close to 0) that solar power will be smaller than load. In this case, both models make perfect predictions too.

Next, let's look into the cases when models fail and find out why the models make wrong predictions. For False Negative (FN) case, we see that solar power prediction is 477.02 KWh, which is slightly lower than the predicted load of 560 KWh. However, real solar generation in this hour turns out to be 784 KWh which is much larger than the predicted value of 477.02 KWh. The classification model also outputs an unsure prediction (close to 0.5) that solar power will be smaller than load. As a result, both models provide the wrong prediction. The reason is that solar power has a dramatic increase of around 42\% in one hour, which could not be forecasted by the models.

For False Positive (FP) case, high contribution of HPow and Irra and low contribution of HLoad output a high predicted solar power and a confident prediction (close to 1) that solar power is larger than load. However, it turns out that both models are wrong. The reason is that solar power generation has a dramatic decrease in this hour. Although the solar power is high in the last hour which is 1012 KWh, it drops sharply to 470 KWh. Besides, the failure of irradiance forecast also contributes to the wrong prediction of two models.

\begin{figure*}[htbp]
    \centering
    \includegraphics [width=0.75\linewidth]{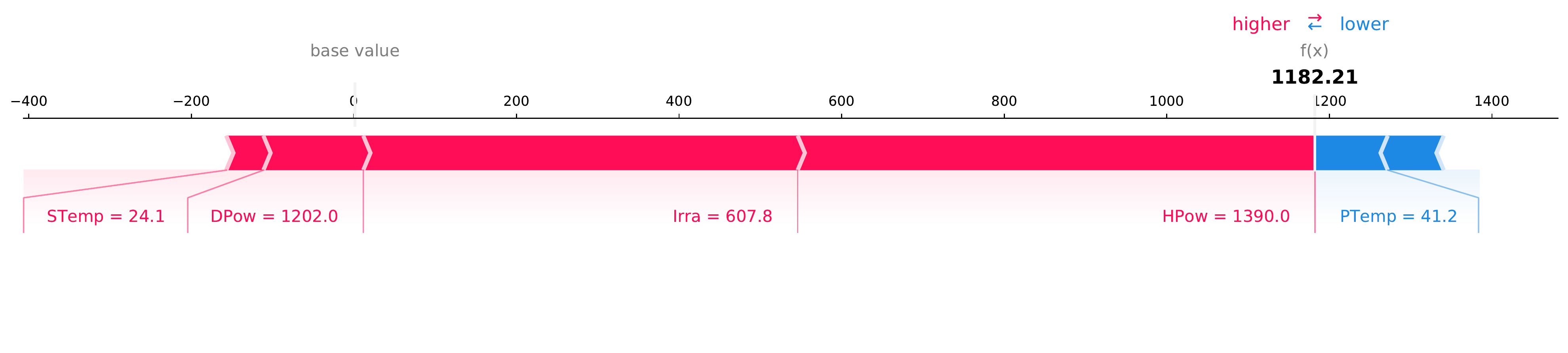}
    \caption{Force plot of the regression model for TP case}
    \label{fig:TP_solar}
\end{figure*}

We use the SHAP tool~\cite{x3} to make a force plot that visualizes the contribution of each feature. A force plot for the regression model of TP case is given in Fig.~\ref{fig:TP_solar}. We see that Irra, HPow, and DPow have large positive contributions and promote the predicted solar power generation to a higher value. On the contrary, high solar panel temperature slightly decreases the solar power output. 

To sum up, we find that the most important features for solar power forecast are HPow (previous hour's generation) and Irra (Irradiance forecast). These two features, together with HLoad (previous hour's load) have significant impacts on the classification model too. If the solar irradiance forecast is accurate and if the solar power does not have a sudden dramatic change within one hour, our models provide accurate predictions. 

\subsection{Uncertainty quantification}
\begin{figure}[htbp]
     \centering
    \includegraphics [width=\linewidth]{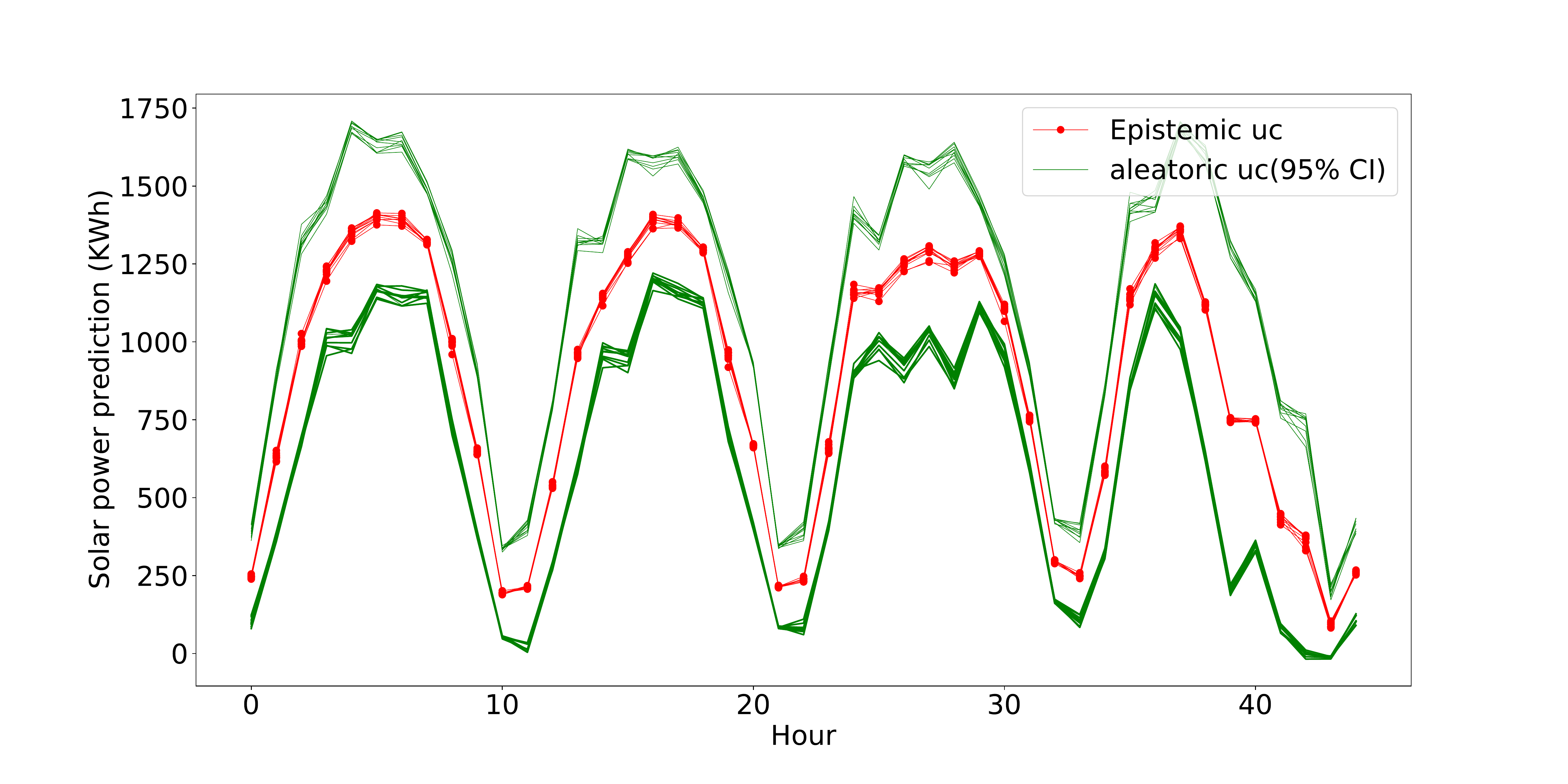}
     \caption{Uncertainty estimation using Bayesian neural networks}
     \label{fig:BNN}
\end{figure}

\begin{figure}[htbp]
     \centering
    \includegraphics [width=\linewidth]{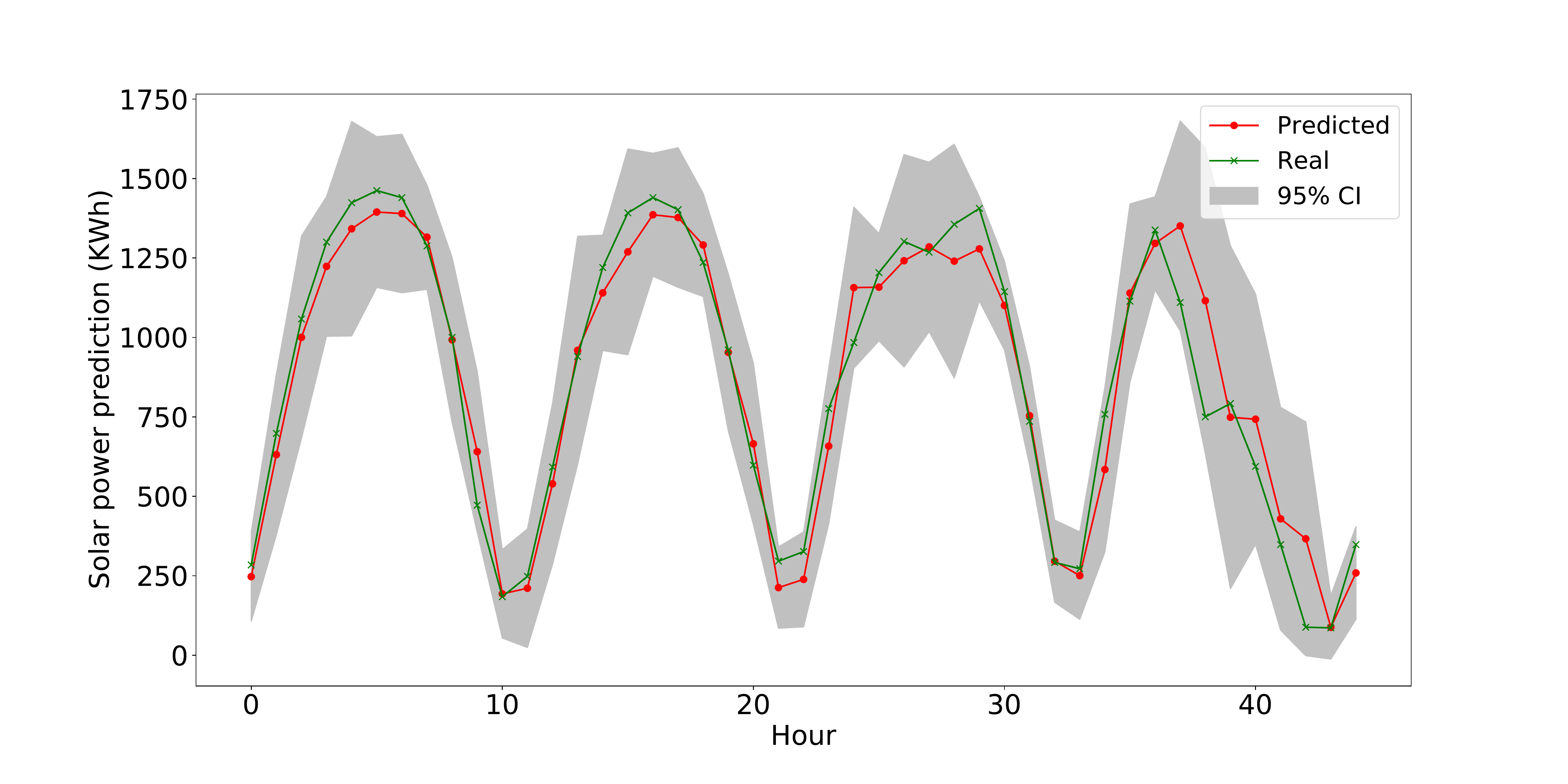}
     \caption{Probabilistic forecast}
     \label{fig:pf}
\end{figure}

The neural networks only provide a point estimation of solar power generation. As a result, the regression neural network in section \ref{sec:interpret} is ``equally confident'' in all described TP, TN, FP, FN cases, while hours during a day are not equivalent in terms of solar irradiance. We propose to address this problem by quantifying the uncertainty of the solar power forecast. Uncertainty estimation of the models allows increasing efficiency of the decision-making process. So, designed neural networks can perform autonomously in hours with low uncertainty. On the contrary, in the hours with high uncertainty, the models can request the intervention of the domain expert. We quantify aleatoric and epistemic uncertainties by designing Bayesian neural networks. Uncertainty estimation using Bayesian neural network is shown in Fig.~\ref{fig:BNN}, where red lines depict different predictions made through Monte Carlo sampling, and the standard deviation of these values can be interpreted as epistemic uncertainty. Green lines represent the standard deviation of the model's output distribution which can be seen as aleatoric uncertainty. We find that both uncertainties are high when solar power generation is high, which happens in the middle of the day. Moreover, aleatoric uncertainty is high in some abnormal hours. Summing up aleatoric and epistemic uncertainties, we plot a probabilistic forecast with 95\% confidence in Fig.~\ref{fig:pf}. While for some abnormal hours, the predicted solar output is far from the real one, it is still within the confidence interval. Bayesian neural networks allow finding the reason for this difference, which is a sudden change in solar power generation. 

\section{Conclusion and future work}\label{sec:conclusion}
In this paper, we consider the problem of designing explainable neural networks, particularly, for solar generation forecasting. Without becoming human interpretable, neural networks cannot be implemented in such critical infrastructure, as power systems. To interpret the designed classification and regression neural networks, we conduct global and local feature analysis. We discover several interesting dependencies between the weather parameters and outputs of the neural networks, which originate from the physical processes during the operation of PV panels. In order to determine the most influential features for forecasting, we implement three state-of-the-art attribution methods: Integrated Gradients, Expected Gradients, and DeepLIFT. We find out that load and solar generation of the previous hour and solar irradiance influence the most to the outcome of the neural networks. In addition, we discover that the main source of the wrong forecasts is the sudden dramatic change of solar irradiance. Such change can be caused, for example, by cloud movements. To resolve this problem we implement Bayesian neural networks, which with the use of historical data, determine hours with high uncertainty of the PV output. As a result, these hours can be known beforehand and considered more carefully by the domain expert. 

We see our work as the first step towards the creation of continuously learning neural networks for PV forecasting. Currently, the designed neural networks can perform autonomously in hours with low uncertainty, while in the hours with high uncertainty, the models can request the intervention of the domain expert. The neural network can learn from the decisions made by the human expert, which will further enhance the neural network performance and make a step towards autonomous forecasting and decision-making algorithms performing on the above human level.

\bibliographystyle{IEEEtran}
\bibliography{references}

\end{document}